%% file: main.tex
\documentclass[conference]{IEEEtran}
\IEEEoverridecommandlockouts
\usepackage[utf8]{inputenc}
\usepackage[T1]{fontenc}
\usepackage{newunicodechar}
\newunicodechar{✓}{\checkmark}
\usepackage{cite}
\usepackage{amsmath,amssymb,amsfonts}
\usepackage{algorithmic}
\usepackage{graphicx}
\usepackage{caption}
\usepackage{subcaption}
\usepackage{textcomp}
\usepackage{xcolor}
\usepackage{textcomp}
\usepackage{anyfontsize}
\usepackage{booktabs}
\usepackage{siunitx}
\usepackage[super]{nth}
\usepackage[para]{threeparttable}
\graphicspath{{./figs/}{.}}

\usepackage{ragged2e}
\usepackage{dirtree}

\usepackage[acronym, style=super, nonumberlist]{glossaries}
\include{acr}

\usepackage{tikz}
\usepackage{pgfplots}
\pgfplotsset{width=7cm,compat=1.8}

\usepackage{url}

\usepackage{fancyhdr}
\fancypagestyle{mahmood}{%
  \fancyhf{} 
  
  \fancyhead[C]{\footnotesize \textcopyright 2021 IEEE. Personal use of this material is permitted.  Permission from IEEE must be obtained for all other uses, in any current or future media, including reprinting/republishing this material for advertising or promotional purposes, creating new collective works, for resale or redistribution to servers or lists, or reuse of any copyrighted component of this work in other works.}
}%
\makeatletter
\let\ps@IEEEtitlepagestyle\ps@mahmood
\makeatother

\begin{document}
\bstctlcite{IEEEexample:BSTcontrol}

\title{
\fontsize{23}{28}\selectfont
SmartHand: Towards Embedded Smart Hands for Prosthetic and Robotic Applications
}
    \author{\IEEEauthorblockN{
    Xiaying Wang\IEEEauthorrefmark{1}\IEEEauthorrefmark{2}, 
    Fabian Geiger\IEEEauthorrefmark{2},
    Vlad Niculescu\IEEEauthorrefmark{1}\IEEEauthorrefmark{2}, 
    Michele Magno\IEEEauthorrefmark{1}\IEEEauthorrefmark{2}, 
    Luca Benini\IEEEauthorrefmark{2}}
    \IEEEauthorblockA{\\[-2mm]\IEEEauthorrefmark{2}ETH Zürich, Dept. EE \& IT,  Switzerland
    }
    \thanks{\IEEEauthorrefmark{1}Corresponding authors. Emails: {xiaywang, vladn}@iis.ee.ethz.ch, michele.magno@pbl.ee.ethz.ch}
    }
    
\maketitle

\begin{abstract}
The sophisticated sense of touch of the human hand significantly contributes to our ability to safely, efficiently, and dexterously manipulate arbitrary objects in our environment.
Robotic and prosthetic devices lack refined tactile feedback from their end-effectors, leading to counterintuitive and complex control strategies. To address this lack, tactile sensors have been designed and developed, but they are either expensive and not scalable or offer an insufficient spatial and temporal resolution. 
This paper focuses on overcoming these issues by designing a smart embedded system, called SmartHand, enabling the acquisition and real-time processing of high-resolution tactile information from a hand-shaped multi-sensor array for prosthetic and robotic applications. 
We acquire a new tactile dataset consisting of 340,000 frames while interacting with 16 objects from everyday life and the empty hand, i.e., a total of 17 classes.
The design of the embedded system minimizes response latency in classification, 
by deploying a small yet accurate convolutional neural network on a high-performance ARM Cortex-M7 microcontroller. 
Compared to related work, our model requires one order of magnitude less memory and 15.6$\times$ fewer computations while achieving similar inter-session accuracy and up to 98.86\% and 99.83\% \mbox{top-1} and \mbox{top-3} cross-validation accuracy, respectively.
Experimental results of the designed prototype show a total power consumption of 505\,mW and a latency of only 100\,ms.

\end{abstract}

    \addtolength{\textfloatsep}{-5mm}
    \addtolength{\dbltextfloatsep}{-3mm}
    \addtolength{\floatsep}{-3mm}
    \addtolength{\dblfloatsep}{-3mm}
    \addtolength{\abovedisplayskip}{-2mm}
    \addtolength{\belowdisplayskip}{-2mm}
    \linespread{0.90}
    \renewcommand{\baselinestretch}{0.90}

\input{sections/introduction}
\input{sections/implementation}
\input{sections/datasets}
\input{sections/results}

\section{Conclusion}
\label{ch:conclusion}

This paper presents SmartHand, a smart embedded system 
that is a step towards equipping robotic and prosthetic devices with a sense of touch by demonstrating the feasibility of acquiring useful tactile data with high spatial and temporal resolution from a highly customizable tactile sensor. Starting from replicating a state-of-art tactile sensor with high spatial resolution, SmartHand focuses on maximizing the temporal resolution by reducing the latency in the system response thanks to the execution of a compact deep learning model for the classification task directly close to the sensor node.
A working prototype of SmartHand is designed and implemented to carry out solid experimental evaluations that have shown a power consumption of 505\,mW and a latency of 100\,ms achieving an inter-session accuracy of 77.84\% in recognizing 16 different objects and the empty hand.

\bibliographystyle{IEEEtran}
\bibliography{bib,myBstCtl}

\end{document}

%% file: acr.tex
\newacronym{adc}{ADC}{analog-to-digital converter}
\newacronym{cnn}{CNN}{convolutional neural network}
\newacronym{cpc}{CPC}{conductive polymer composite}
\newacronym{dma}{DMA}{direct memory access}
\newacronym{eth}{ETH}{Eidgenössische Technische Hochschule}
\newacronym{fsr}{FSR}{force sensitive resistor}
\newacronym{fps}{FPS}{frames per second}
\newacronym{gpu}{GPU}{graphics processing unit}
\newacronym{gui}{GUI}{graphical user interface}
\newacronym{i2c}{I2C}{inter integrated circuit}
\newacronym{ic}{IC}{integrated circuit}
\newacronym{iis}{IIS}{Integrated Systems Laboratory}
\newacronym{imu}{IMU}{inertial measurement unit}
\newacronym{led}{LED}{light-emitting diode}
\newacronym{ldo}{LDO}{low-dropout}
\newacronym{macc}{MACC}{multiply-and-accumulate operation}
\newacronym{mcu}{MCU}{microcontroller unit}
\newacronym{mems}{MEMS}{microelectromechanical system}
\newacronym{mit}{MIT}{Massachusetts Institute of Technology}
\newacronym{mlp}{MLP}{multi-layer perceptron}
\newacronym{pcb}{PCB}{printed circuit board}
\newacronym{pulp}{PULP}{parallel ultra low power}
\newacronym{ram}{RAM}{random-access memory}
\newacronym{sdram}{SDRAM}{synchronous dynamic random-access memory}
\newacronym{spdt}{SPDT}{single pole double throw}
\newacronym{spi}{SPI}{serial peripheral interface}
\newacronym{stag}{STAG}{scalable tactile glove}
\newacronym{tcn}{TCN}{temporal convolutional neural network}
\newacronym{uart}{UART}{universal asynchronous receiver-transmitter}
\newacronym{wom}{WOM}{wake-on-motion}
\newacronym{afe}{AFE}{analog front-end}
\newacronym{cv}{CV}{cross-validation}

%% file: sections/introduction.tex
\section{Introduction}
\label{ch:introduction}

Robotic hands and prosthetic devices provide robots with humans' ability to interact with the surroundings and restore lost abilities for amputees. 
The preeminent modality of choice for robotic devices that need to interact with the environment is vision~\cite{BERGAMINI2020101052,Cheng2020visiontim}. 
Optical sensors are performing well, are inexpensive, and mature computer vision algorithms exist for every level of processing power.
On the other hand, there are tasks in which optical information is insufficient or leads to disproportionately complex control strategies.
One such task is the manipulation of arbitrary objects with an articulated end-effector. Vision is required to find correctly the object and the position of the end-effector, but once contact is made, the robot cannot tell if the object is fragile or robust, if the end-effector has a good grip or if the object will slip from its grasp. For this reason, a recent trend is to develop non-visual technologies and methods to provide a robot with those capabilities.

In nature, organisms have evolved a useful modality for manipulating their surroundings: the touch.
Science is often inspired by nature, so in order to improve the ability of robots and prosthetic devices to manipulate objects in the environment, tactile sensors have recently gained considerable attention.
Even though much research has been conducted~\cite{weiner2020embeddedfingers, Zou2017, Kappassov2015, Xu2018, maddipatla2017pressuresensor}, the task of developing a tactile sensor with a resolution comparable to the human hand, with similar robustness as well as flexibility, remains an open challenge~\cite{Sundaram2019}.
Object grasping and manipulation are natural abilities for healthy human beings, but it is a very difficult task for robotic machines. Besides the sense of touch, humans can elaborate sensory and perceptive inputs during object interaction. The inertial properties of the object are immediately perceived, and the corresponding adjustments are performed in real-time during the manipulation.
This is an essential ability that a robotic machine has to be equipped with in order to optimally interact with objects. In fact, the weight, the mass distribution, and the inertial resistance are all factors to be considered for efficient and safe robotic motor planning. 
However, no satisfying solution has yet been proposed due to the lack of well-defined sensing equipment that provides useful data for estimating the inertial parameters~\cite{MAVRAKIS2020inertial}. 

State-of-the-art sensor solutions, from academia and industry, are typically hard to manufacture, expensive, and lacking in temporal as well as spatial resolution~\cite{Yin2018, Zhang2018, Franceschi2017}. The majority of the solutions mainly focus on replicating the fingers tips~\cite{syntouch,weiner2020embeddedfingers,Choi2020finger} or have sensors covering only a limited part of the hand palm~\cite{Kang2017touchsensor}.
Recently, a remarkable approach has been proposed by Sundaram and colleagues\cite{Sundaram2019}. For the first time in literature, the authors have created a \gls{stag} that presents the full coverage of the human hand, providing tactile information with high spatial resolution.
A tactile dataset is collected while interacting with 26 daily life objects for several minutes each. A \gls{cnn} is subsequently applied for classifying the objects and the empty hand.
To be noted that the proposed setup with offline sensor data processing based on several-minute recordings introduces long latency, making it sub-optimal for a real-world application where a prompt response in the range of milliseconds is desired. 
Moreover, embedding multi-sensors in small integrated systems is actively researched thanks to its advantages for closed-loop control of the robotic devices~\cite{Yunas2020sas}. Weiner and colleagues~\cite{weiner2020embeddedfingers} propose a fingertip system integrating several sensors that are miniaturized to fit in the very limited space of a fingertip. Besides the integration of sensors, a smart processing unit, e.g., \glspl{mcu}, that can process the acquired sensor data at the edge would enable a real-time response of the whole system.

Recent developments in edge processing units and tinyML~\cite{wang2020fann} make it possible to bring the data processing close to the sensors. Along with efforts in designing and deploying tiny machine learning and deep learning models on the edge, nowadays edge devices are becoming increasingly smart and can interpret sensor data and translate it into actionable information in real-time~\cite{scherer2021radar}. 
A drawback of low-power edge computing is that the resources available on \glspl{mcu} with milli-watt power consumption are very limited making it impossible to deploy big and complex models. Thus, a conscious design of tiny yet accurate models is necessary, by taking into consideration the memory availability and computational capability of the underlying processor.

This work is a step towards equipping robotic and prosthetic devices with a smart sense of touch by presenting a smart embedded system, called Smarthand, capable of reading and processing high-resolution tactile information in real-time. 

The main contributions of this paper are as follows:
\begin{itemize}
    \item We reproduce a multi-array tactile sensor in the shape of a human hand with high spatial resolution~\cite{Sundaram2019} and integrate it in a full system, called SmartHand, able to acquire tactile and hand movement data and process it in real-time.
    \item Three operational modalities for different use-cases were implemented, and a large dataset was collected while interacting with 16 objects from everyday life in addition to the empty hand. The dataset is open-source released\footnote{Available at \url{https://iis.ee.ethz.ch/~datasets/smarthand}.}.
    \item We design a novel \gls{cnn} which requires one order of magnitude less memory footprint and 15.6$\times$ less computation with negligible accuracy drop of 0.21\% when eight input images are given or around 2\% when a single frame is given, compared to the original network on the STAG dataset from MIT~\cite{Sundaram2019}.
    \item We successfully deployed the trained model on an ARM Cortex-M7 \gls{mcu} close to the sensor node, hosted by the SmartHand, to enable smart edge computing. The \mbox{top-1} and \mbox{top-3} classification accuracy is respectively 98.86\% and 99.83\% when seven-fold cross-validation is performed on the full acquired dataset. While the inter-session \mbox{top-1} and \mbox{top-3} accuracy values are respectively 49.63\% and 77.84\% with only one input frame, which significantly increases the temporal resolution by reducing the system response from several minutes to sub-second.
    \item Experimental measurements show that our proposed SmartHand, with embedded \gls{cnn}, can infer the contacted object with a low latency of 100\,ms while consuming a total power of 505\,mW.
\end{itemize}

%% file: sections/implementation.tex
\section{System Architecture} \label{ch:sysarch}

\begin{figure}[t]
  \centering
  \includegraphics[width=.95\linewidth]{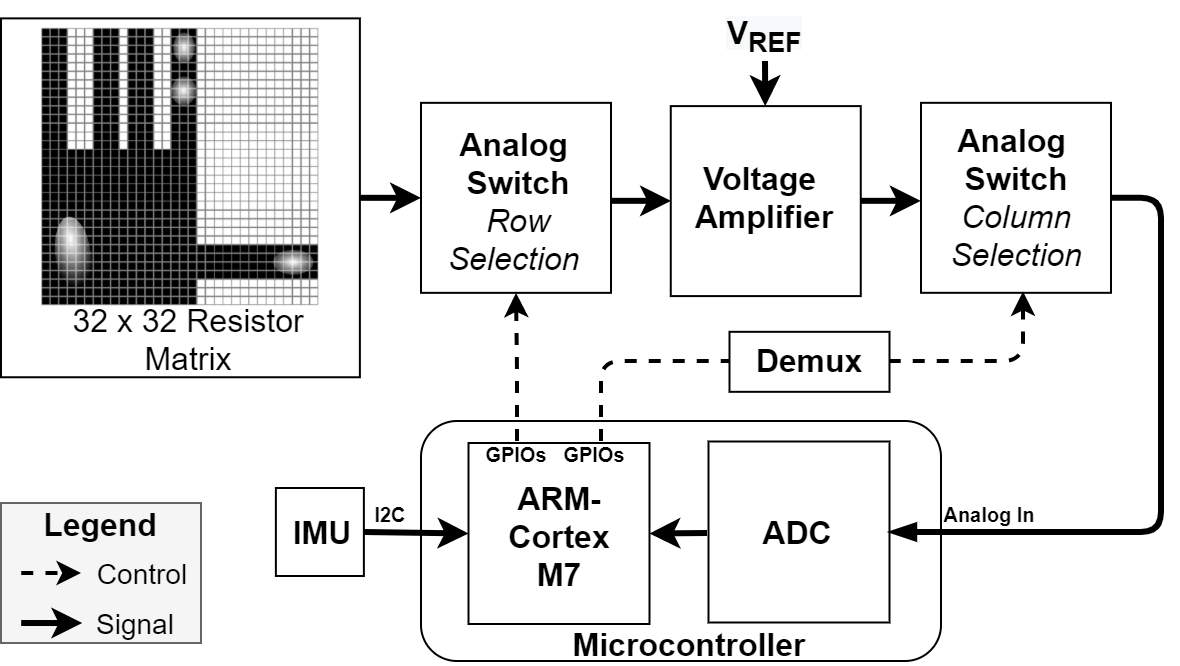}
  \caption{The block diagram of the system.}%
  \label{fig:block}
\end{figure}

\begin{figure}[t]
  \centering
  \includegraphics[trim={0cm 10cm 13cm 6cm},clip, width=.8\linewidth]{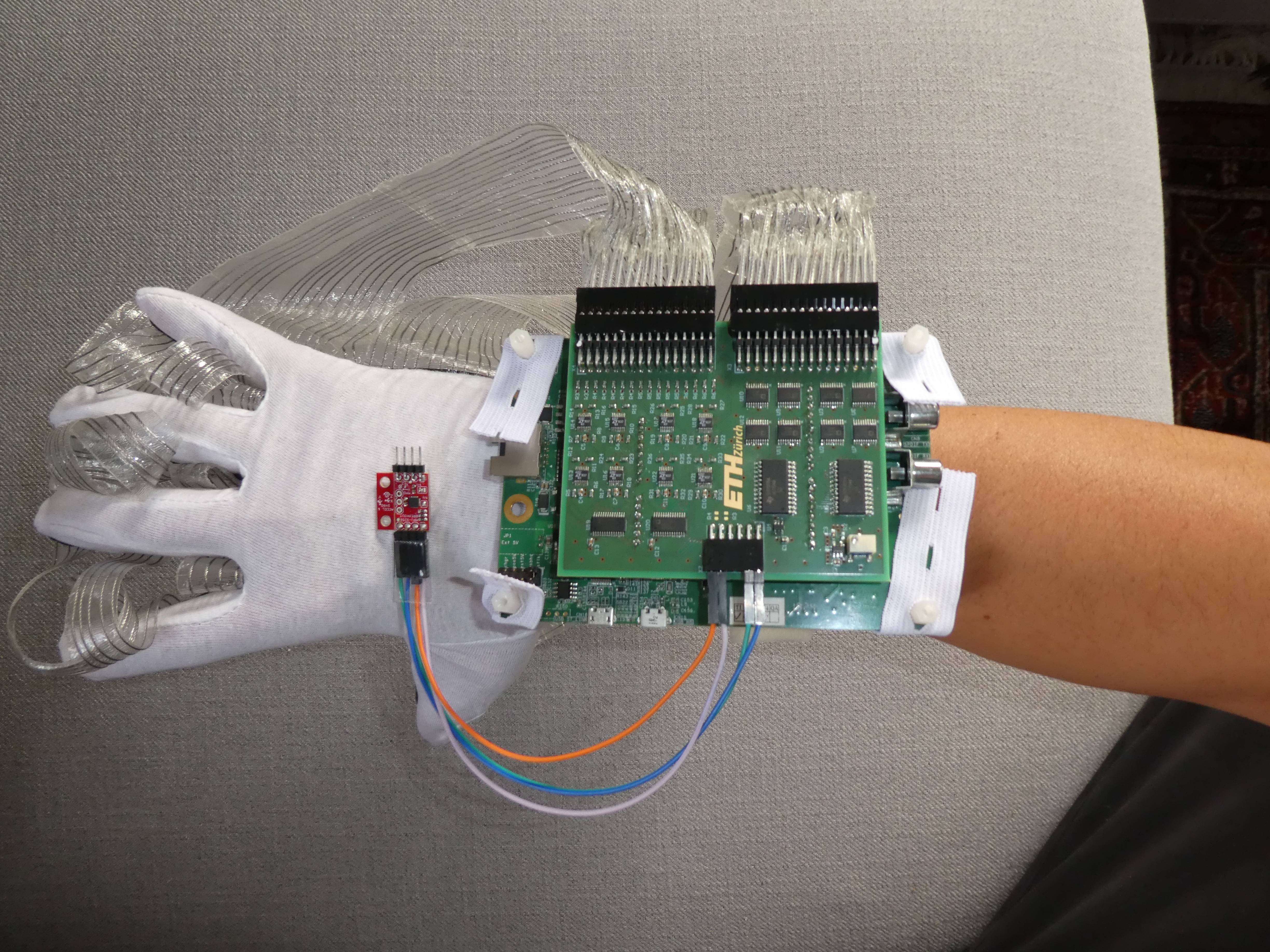}
  \caption{Fully assembled system from the top. The red PCB on the back of the glove is the IMU. The readout circuit is plugged onto the discovery board like a shield.}%
  \label{fig:full_system_top}
\end{figure}

\begin{figure}[t]
  \centering
  \includegraphics[trim={5cm 10cm 5cm 0},clip, width=.8\linewidth]{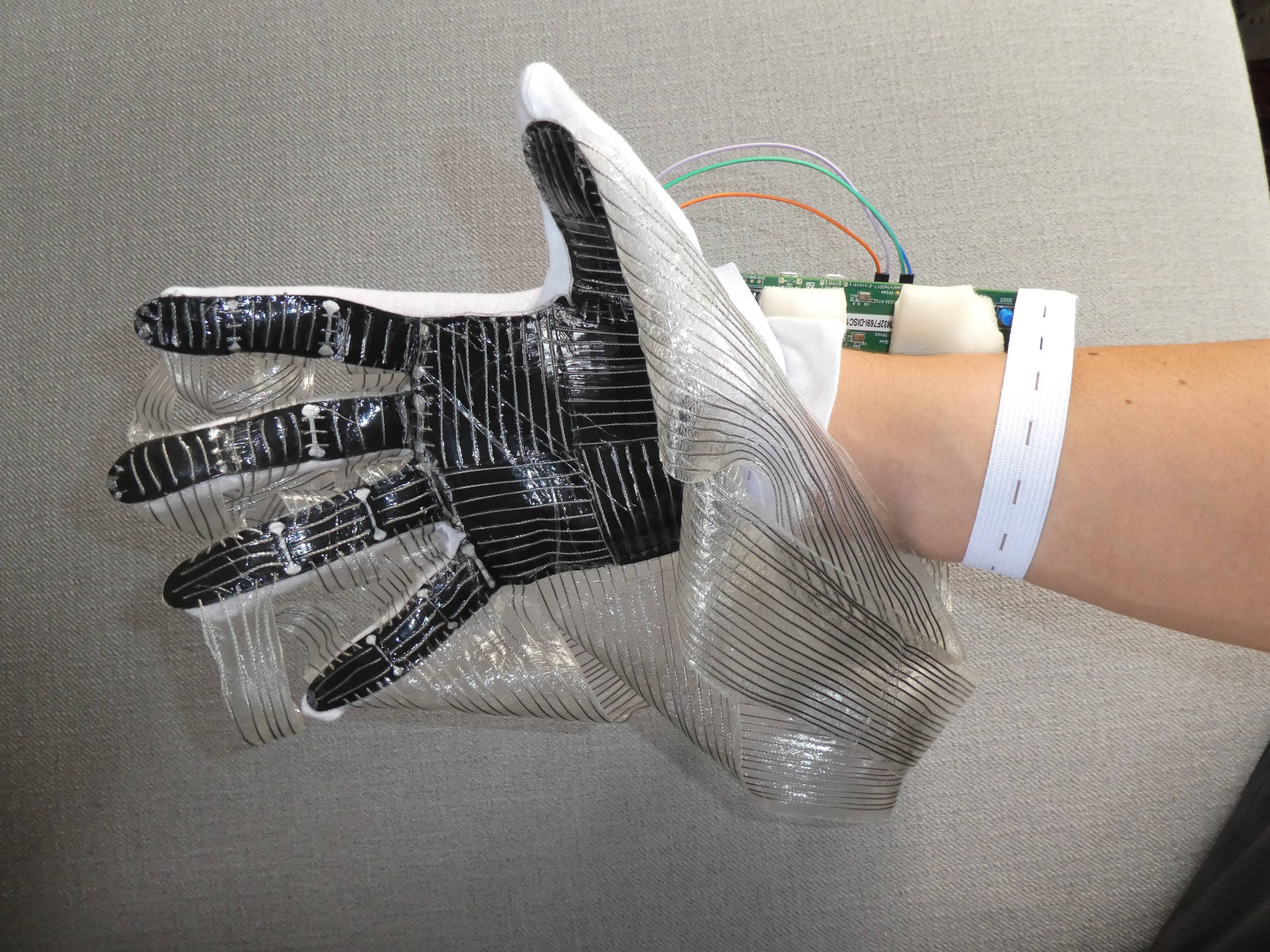}
  \caption{Fully assembled system from the bottom. The sensor laminate is glued to the white cotton glove.}%
  \label{fig:full_system_bot}
\end{figure}

Fig.~\ref{fig:block} depicts the block diagram and the architecture of the designed Smarthand. It consists of a low cost resistive  tactile sensor based on a \gls{cpc} as in~\cite{Sundaram2019}, an \gls{afe} to read the tactile data, an \gls{imu} to gain additional movement information, and an \gls{mcu} for system control and onboard signal processing.
Fig.~\ref{fig:full_system_top} and Fig.~\ref{fig:full_system_bot} show the full system assembled and worn.


\subsection{Tactile sensor}
\begin{figure}[t]
  \centering
  \includegraphics[width=.65\linewidth]{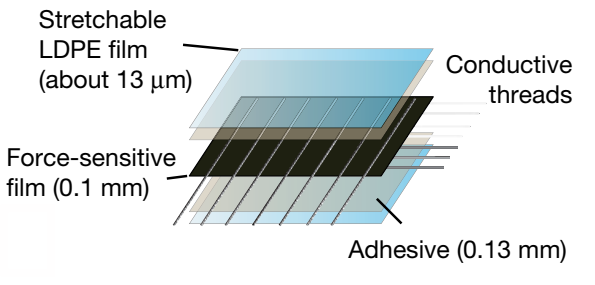}
  \caption{Sensor composition. Adhesive and LDPE film are required to bond the electrodes to the force sensitive film and protect it, respectively~\cite{Sundaram2019}.}%
  \label{fig:sensor_structure}
\end{figure}

\begin{figure}[t]
  \centering
  \includegraphics[width=.75\linewidth]{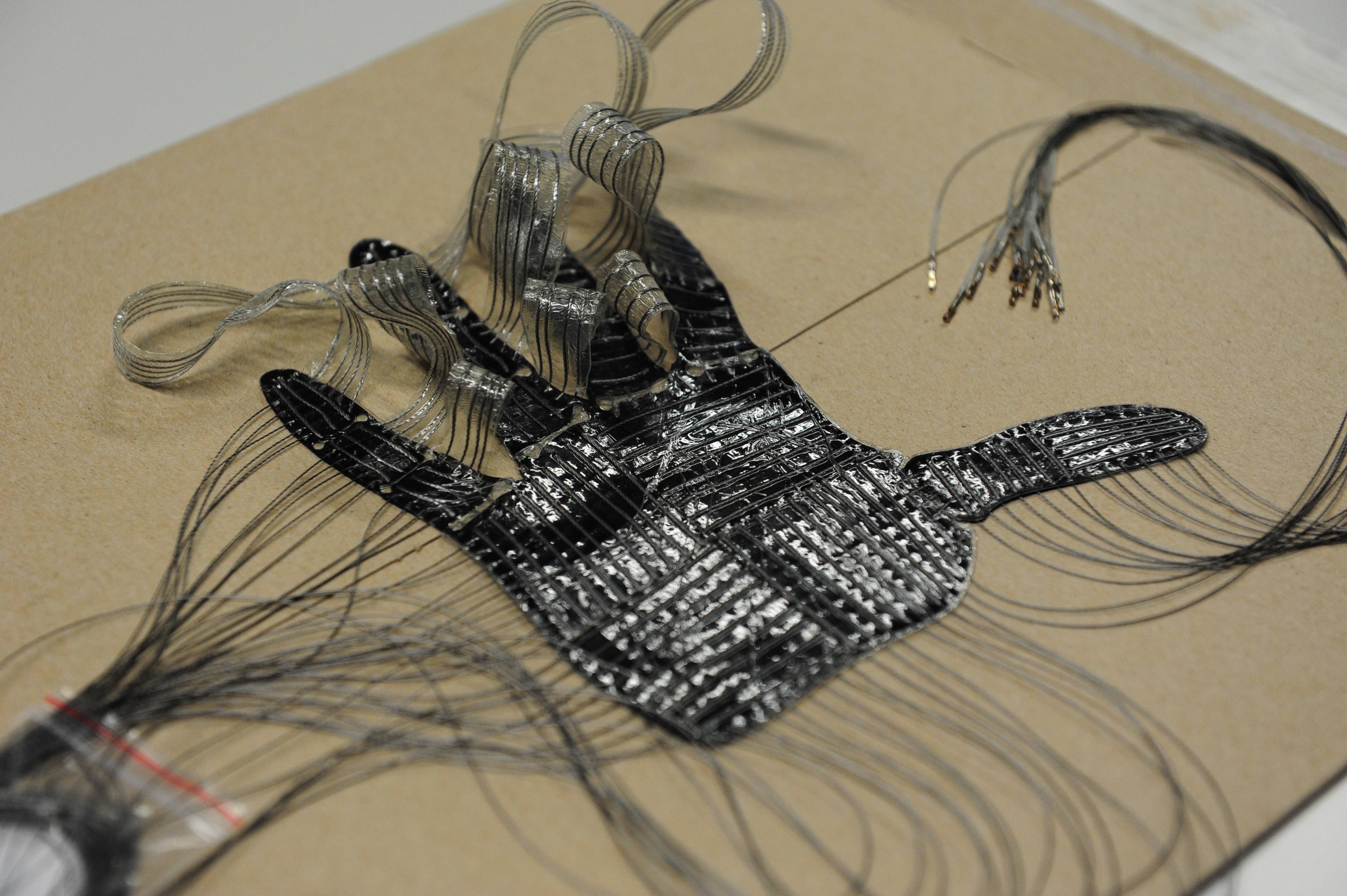}
  \caption{Fabricated sensor before being glued to a glove.}%
  \label{fig:sensor_thesis}
\end{figure}

\begin{figure}[t]
  \centering
  \includegraphics[width=.75\linewidth]{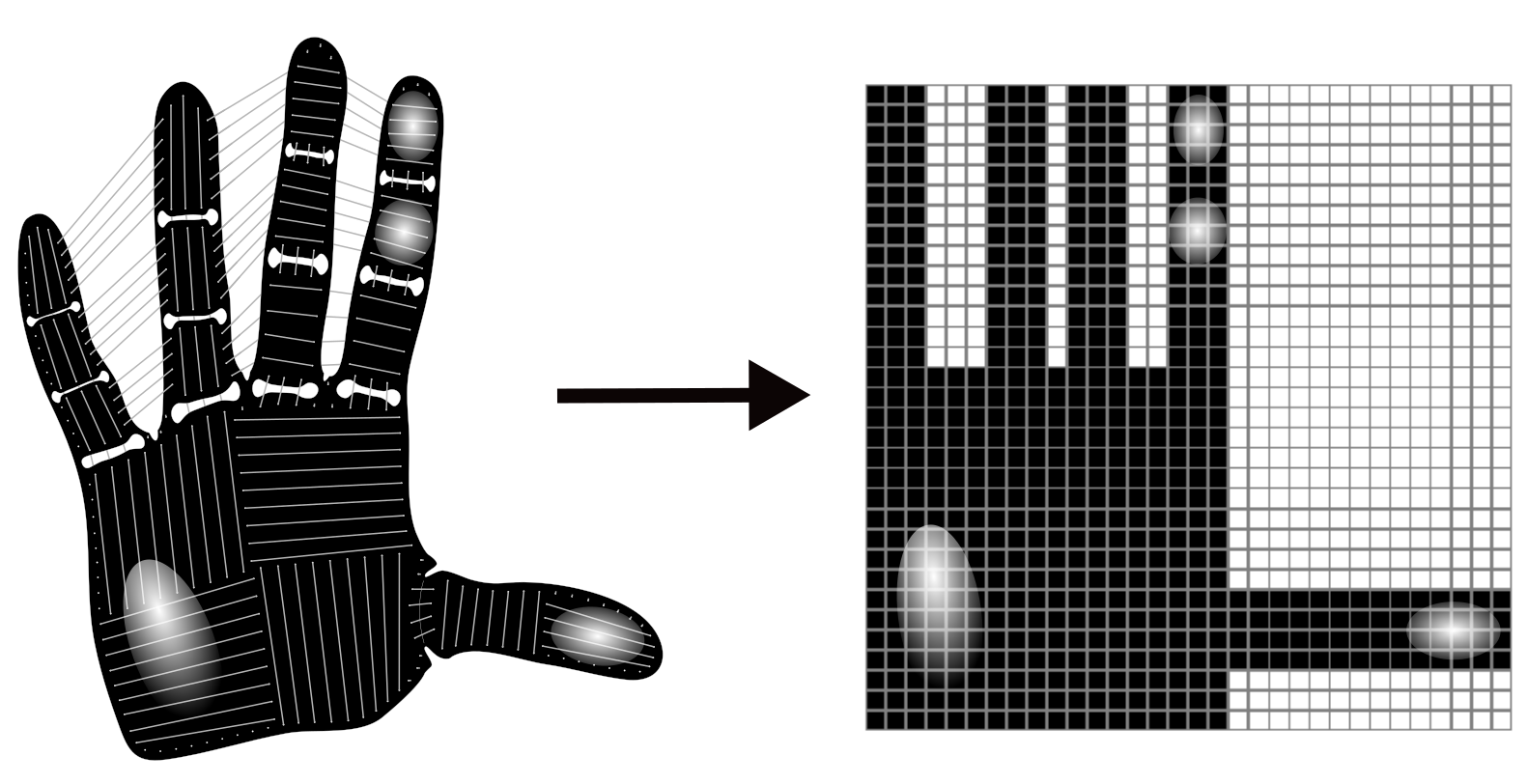}
  \caption{Representation of the sensor values as a 32x32 matrix, referred to as \emph{frame} or \emph{tactile frame}.}%
  \label{fig:glove2frame}
\end{figure}

Following the fabrication procedures described by~\cite{Sundaram2019}, a tactile sensor with high spatial resolution was produced in our laboratory. 
The sensor consists of a \gls{cpc}, called Velostat, sandwiched between two orthogonal sets of electrodes, as shown in Fig.~\ref{fig:sensor_structure}. Based on the concept of \gls{fsr}, in response to increasing normal forces, its resistance decreases, which can be read with a dedicated analog front-end.
There are 32 electrodes running across each of the two sides of the force sensitive film forming a grid.
The function of the sensor does not depend on its shape, thus it can be fabricated in arbitrary forms and sizes.
To obtain comparable results and capture useful data during object interaction, the sensor was fabricated as similarly as possible to~\cite{Sundaram2019}. Fig.~\ref{fig:sensor_thesis} shows the finished sensor laminate before being attached to a thin cotton glove.
Because the active sensor area is shaped like a hand, there are 548 physical crossings of electrodes.
However, to preserve the spatial relation between taxels, or sensor points, and make it easier to work with the sensor data, all 1024 available taxels are read.
By arranging the read values in a 32x32 matrix, an intuitive and natural representation of the sensor can be constructed, as shown in Fig.~\ref{fig:glove2frame}, and will be referred to as \emph{tactile frame}, or simply \emph{frame}.

\subsection{Readout Circuit}
\label{subsec:readout_circuit}

The authors in~\cite{Sundaram2019} designed a readout circuit for their \gls{stag}. Abstracting from its physical structure, the tactile sensor is a network of resistors formed by the cross-talks between the two orthogonal sets of 32 electrodes. Each cross-talk can be seen as a simple resistor with variable resistances depending on the pressure applied to the sensor glove. The readout circuit has to be able to read each of these resistance values correctly. The circuit proposed by the authors is based on the electrical isolation scheme~\cite{Kim2016}, which ensures that there is only one current path when measuring the resistance between two electrodes.  
Our implementation is based on this original design but improved by using a more powerful \gls{mcu} featuring a higher precision analog-to-digital converter (12-bit).

\subsection{Movement sensor}
We additionally integrate an \gls{imu} for acquiring accelerometer and gyroscope data to collect hand movements during object interaction.
The used \gls{imu} is \mbox{MPU-9250} with a sampling frequency of \SI{100}{\hertz} and a resolution of 16-bit for both the accelerometer and the gyroscope.

\subsection{Edge processing unit}
The embedded system requires an \gls{mcu} with the ability to drive and read the tactile sensor, as well as directly run neural network inferences with the acquired data and take action based on the inference, e.g., actuating the motors of the robotic arm.
For a fast development of the prototype, the STM32F769NI discovery board was chosen, featuring an ARM Cortex-M7 core. With its 2\,MB of Flash memory, 532\,kB of RAM, and a maximum clock speed of 216\,MHz, it is one of the most highly-performing \gls{mcu} in the low-power ARM Cortex-M family.
It comes with the software tool STM32CubeMX, from STMicroelectronics, employed for generating initialization code, compiling, and flashing.
\mbox{X-CUBE-AI} is an expansion package for STM32CubeMx that allows fast deployment of neural networks on STM32 \glspl{mcu}.

\subsection{Operational Modalities}\label{subsec:data_collection}
The firmware of the system is implemented such that three operational modalities are available for different use-case scenarios. A basic timer peripheral on the \gls{mcu} is used to clock the applications.
\subsubsection{Data collection}
This modality is used for acquiring a complete dataset. The system waits for a `start' command, given either by pressing a button on the discovery board or sending the character `r' via the \gls{uart} protocol to the \gls{mcu}. 
If either of these two events is detected, the base timer is triggered at a rate of 100\,Hz, and the tactile along with IMU data is collected.
The external \gls{sdram} on the discovery board was utilized as intermediate storage for tactile data, while the \gls{imu} data is sent directly after its acquisition. 
After a configurable number of frames has been collected, the data acquisition stops and immediately sends the content of the \gls{sdram} to a connected computer, and the system returns to the idle state waiting for the next `start' command.
The \acrshort{sdram} can store a maximum of 4096 tactile frames, limiting one single interaction to a maximum of about 40 seconds with a data collection rate of 100Hz. The acquisition frequency can be reduced if a longer recording is desired.

\subsubsection{Real-time data visualization}
A \gls{gui} designed using Python is implemented and can be used to visualize the tactile sensor data on a computer screen directly. Once the `start' command is received, the base timer is activated, and the tactile frames are collected at 10\,Hz and directly sent to the connected computer via \acrshort{uart}. To stop the visualization, either the button on the discovery board has to be pressed again, or the character `p' needs to be sent to the system.
The \gls{gui} displays each frame as a 32x32 matrix and provides two buttons, labeled `Run' and `Pause', which respectively send `r' or `p' to the system, giving the option to control the data visualization from the same interface.

\subsubsection{Smarthand system}
\begin{figure}
  \centering
  \includegraphics[trim={11cm 0 14cm 0},clip, width=.85\linewidth]{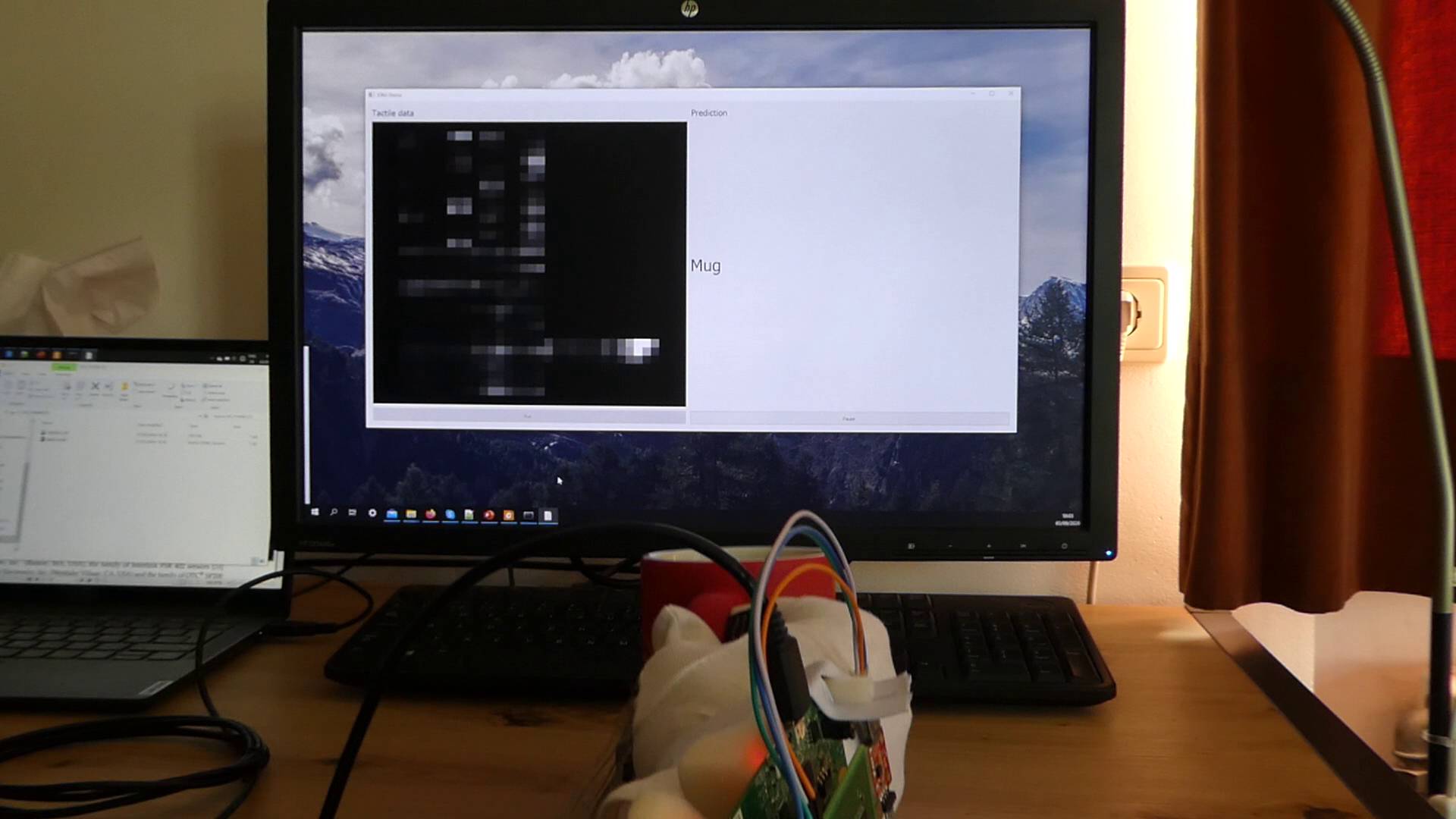}
  \caption{Snapshot from a demo video in full-system modality.}%
  \label{fig:demo_snapshot}
\end{figure}

This modality leverages the full system by not only reading tactile data but also processing it real-time on the \gls{mcu} using a deployed neural network. 
The data acquisition starts after pressing the button on the discovery board with the base timer configured to 8\,Hz. The sensor data is read and processed by the neural network on the \gls{mcu}. The system clock is configured to 216MHz to maximize the inference speed.
To evaluate the performance of the SmartHand system, a second Python \gls{gui} is developed, which displays the result of each network inference and its corresponding input frame. To simultaneously process the frame onboard and visualize it in real-time, a \gls{dma} controller is used for sensor data transmission. It is to note that in this case, the connection to the computer serves merely for the verification purpose of a demo, the system can fully function independent of any processing engine other than the onboard \gls{mcu}, and the output of the neural network inference can be used in real-time for any robotic or prosthetic control. 
Fig.~\ref{fig:demo_snapshot} displays a snapshot of a demo video in this application modality. 
The SmartHand is worn like a glove, and the tactile data during the interaction with a mug is acquired and classified immediately with the onboard embedded \gls{cnn}. The classification output is displayed in real-time on the screen.

\section{Neural Network} \label{sec:cnn}
This section describes the model design, the training and the evaluation of the neural network used for processing the sensor data, and its embedded deployment on the onboard \gls{mcu}.

\subsection{Convolutional Neural Network Design}
\begin{figure*}
  \centering
  \includegraphics[width=.85\linewidth]{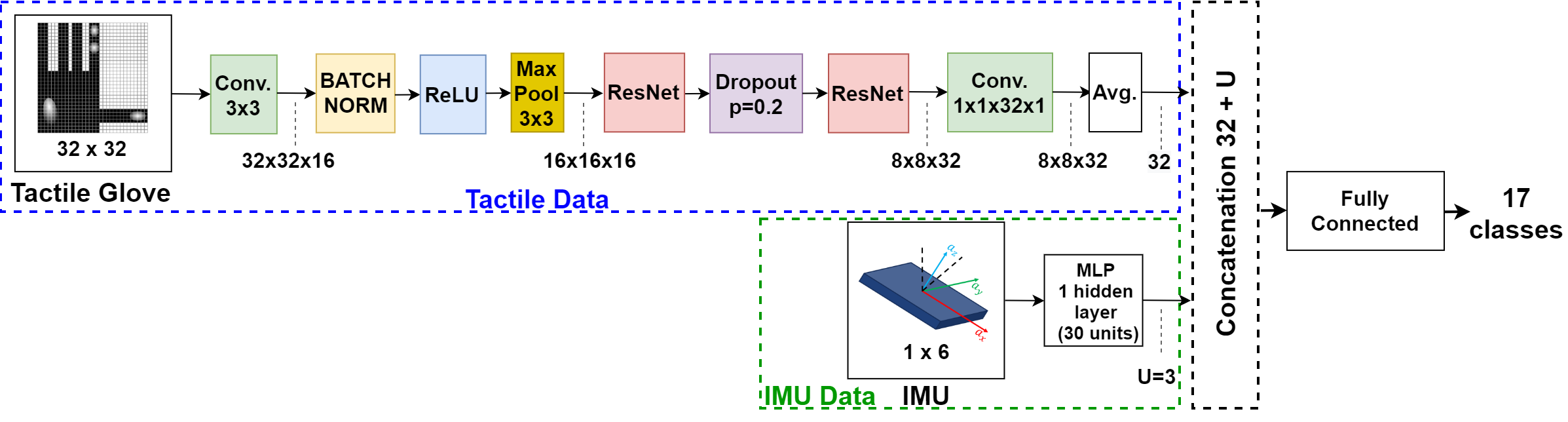}
  \caption{The proposed \gls{cnn}.}
  \label{fig:adapted_cnn}
\end{figure*}

Because of the inherently spatial information present in the sensor, a \gls{cnn} architecture was chosen to process the tactile data. 
As in~\cite{Sundaram2019}, we choose a model architecture based on \mbox{ResNet-18}~\cite{he2016resnet18}, but fully redesign it by taking into consideration hardware resource constraints. Fig.~\ref{fig:adapted_cnn} depicts the proposed model architecture framed with a blue dashed line. 
It uses multiple stages of convolutions, including two residual blocks, to extract features from a single input frame. Finally, a fully connected layer is used for the classification.
Compared to~\cite{Sundaram2019}, we restrict the number of input frames to a single one directly reducing the latency.
Another adaptation is a four-fold reduction of convolution filters. This decreases significantly the inference speed thanks to the reduced model complexity.

To include the \gls{imu} data, a simple \gls{mlp} with one hidden layer consisting of 30 hidden units is added to extract the feature representations. The output layer of the \gls{mlp} consists of 3 neurons and is subsequently concatenated to the features of the tactile data before the final fully connected layer for the classification. This method is motivated by literature~\cite{Yunas2020sas,Trumble, Li2019}, in which the features of the individual sensors are first extracted and then concatenated. This additional branch featuring \gls{imu} data can be added or removed depending on the application.

\subsection{Training and validation}
The model is trained using PyTorch on GeForce GTX 1080 Ti GPUs with CUDA framework. 
Two validation methodologies were employed. First, the data is randomly split, and seven-fold \gls{cv} is performed. 
Second, the data is split by session to consider the inter-session variability. One session is kept as a validation set, while the remaining four are used for training.
In both cases the model is trained for 60 epochs with cross-entropy loss and Adam optimizer.

\subsection{Emdedded implementation}
The X-CUBE-AI expansion package for STM32CubeMx is used for deploying the trained \gls{cnn} on the \gls{mcu}. For the embedded deployment, the model is trained by leveraging the full dataset. It is then saved in ONNX format and imported in STM32CubeMx. The tool generates an application template code, to which the user code to read the data can be added and the inference performed.

%% file: sections/datasets.tex
\section{Datasets} \label{sec:datasets}

Two datasets are evaluated in this paper. In order to reproduce the results in~\cite{Sundaram2019} and to fairly compare our proposed methods, we use the dataset provided by the authors. 
With the goal of integrating a complete system capable of sensor data acquisition and online data processing, we acquired our dataset using the in-house produced tactile sensor glove.

\subsection{MIT-STAG Dataset}

The work by Sundaram and colleagues~\cite{Sundaram2019} from MIT 
proposed for the first time a tactile glove covering the full hand, called \gls{stag}, based on the concept of \gls{fsr} using a \gls{cpc}, with high spatial resolution. 
A total of 135,000 frames of tactile data was collected together with synchronized video footage in three recording sessions, during each of which the sensor glove was worn throughout the entire session.
This dataset contains several minutes of interaction with 26 different objects from everyday life. 
Tactile frames were collected at a rate of about 7.3\,Hz. All objects were manipulated in a repeatable way, mostly pushing the sensor from the top onto the object lying on a table.
The frames corresponding to the actual touch of the object are chosen by comparing the values with a threshold found from an empty-hand dataset.
This threshold value is found for each taxel to account for any internal stress caused by the sensor conformation. 
The authors select N frames from the several-minute recordings by either random selection or clustering to construct the input samples to the neural network.
The dataset is publicly available\footnote{http://stag.csail.mit.edu/}.

\subsection{ETHZ-STAG Dataset}

Following the analogous procedures of fabrication, we created our own \gls{stag} at ETH Zurich. 
A large dataset was collected with the \emph{data collection} modality described in Sec.~\ref{subsec:data_collection} in five recording sessions. Instead of synchronous video footage, synchronous accelerometer and gyroscope data are collected from an \gls{imu}.
During every recording session, 16 different objects were manipulated for 40 seconds each while continuously collecting tactile and \gls{imu} data at a rate of 100\,Hz. 
The manipulations were done in the same manner as in~\cite{Sundaram2019} to increase repeatability between recording sessions.
The valid frames of actual object contact are marked by comparing to a threshold value for each taxel, as in~\cite{Sundaram2019}.
We construct these thresholds by finding the highest possible value that a taxel can assume using more than 20,000 additional empty-hand-frames of different hand poses without object contact.
Finally, a total of 340,000 frames are available.

%% file: sections/results.tex
\section{Results}
\label{ch:results}

We present the results of object classification using our proposed \gls{cnn} and compare it to previous work~\cite{Sundaram2019}. In addition, we report the outcome of the embedded implementation in terms of \glspl{macc}, memory usage, and inference time on the selected \gls{mcu}.

\subsection{Neural network performance}

\input{fig/accuracy_res}

First of all, we compare the accuracy of our proposed model with the original network on MIT-STAG dataset~\cite{Sundaram2019}. The authors randomly select N frames from each recording as the input to the \gls{cnn}. We reproduce their results and train our model on the same dataset with the same splitting. As shown in Fig.~\ref{fig:comparison_acc}, despite the significantly smaller size of our model, the accuracy is comparable, especially for N = 8, where the accuracy drop is only 0.21\%. However, N = 1 would introduce less computational burden for a real-time application scenario yielding lower latency. Again, our model performs comparably to the original one, with less than 2\% accuracy drop with \mbox{N = 1}.

The second step is to evaluate the model performance on our dataset collected by the \gls{stag} fabricated in-house and integrated into our proposed system.
With our dataset, we consider only N = 1 and take all the valid frames instead of randomly selecting subsets of valid frames from each recording. This reduces the several-minute latency of previous work and enables the real-time response of the entire system. 
Fig.~\ref{fig:randomsplit} shows the results for the seven-fold cross-validation with random splitting on the whole dataset without considering inter-session variability.
The top-1 and top-3 accuracy values, i.e., the correct class is the one with the highest predicted probability, and the correct class is among the ones with the three highest probabilities, are respectively 98.86\% and 99.83, demonstrating the effectiveness of the neural network to learn the full data distribution.
Whereas, by considering the inter-session variability, we obtain the top-1 and the top-3 accuracy of 49.63\% and 77.84\%, respectively. 

\subsection{Embedded implementation}

We subsequently proceed with network deployment on the \gls{mcu}. Here the main advantage of our proposed model becomes evident.
In fact, the proposed network in~\cite{Sundaram2019} is too big for the selected \gls{mcu} making it impossible to be embedded onboard. Table~\ref{tab:network_comparison} reports the comparison between the two networks. 
Note that the inference time reported for the original network is obtained by reducing the input convolution filters from 64 to 54 since it is impossible to deploy the full network due to memory constraints.
Our model requires one order of magnitude less memory, making it possible to be deployed. The number of computations is reduced by 15.6$\times$ in terms of \glspl{macc}, yielding more than 12$\times$ speedup for the inference time. This effectively enables a system with real-time response. 

\begin{table}[b]
  \caption{Comparison of neural networks.}
  \label{tab:network_comparison}
  \centering\begin{tabular}{@{}lrrrr@{}} \toprule
  \textbf{Project} & \textbf{\acrshort{macc}} & \textbf{Flash} & \textbf{\acrshort{ram}} & \textbf{Inference time}\\ \midrule
  Cit.~\cite{Sundaram2019} & 73.3 M & 2.77 MB & 196 kB & >1.2 s \\
  This work & 4.7 M & 177 kB & 52 kB & 100 ms \\ \bottomrule
  \end{tabular}
\end{table}

\subsection{Sensor fusion}

The \gls{imu} orientation in the form of Euler angles is directly fed into the \gls{mlp}, or only accelerometer or gyroscope data is used.
Unfortunately, no configuration has provided an improvement over the model using only tactile data.
The usage of \gls{imu} data in this application scenario is not very meaningful, as all the objects were mostly manipulated from the top.
In a real application scenario with robotic hands, the information about end-effector orientation is crucial and needs to be included at various stages of the processing loop.
Accordingly, the integration of the \gls{imu} is a step towards a usable and practical system.

\subsection{Power measurements}

Finally, we measure the power consumption of the full system with the \gls{imu} and the \gls{mcu} supplied at 3.3\,V, and the readout circuit at 5V. The measured active power is 505\,mW, and the idle power is 185\,\textmu W.

%% file: fig/accuracy_res.tex
\definecolor{redisch}{RGB}{211,94,96}
\definecolor{greyisch}{RGB}{128,133,133}

\begin{figure}
\centering
  \resizebox{0.85\columnwidth}{!}{%
    \centering

\begin{tikzpicture}
\begin{axis}[
height=5cm,
width=9cm,
legend cell align={left},
legend style={at={(0.97,0.03)}, anchor=south east, draw=white!80.0!black},
tick align=outside,
tick pos=left,
x grid style={white!69.01960784313725!black},
xlabel={Number of channels},
xmajorgrids,
xmin=0, xmax=9,
xtick style={color=black},
xtick = {0, 1, ..., 9},
y grid style={white!69.01960784313725!black},
ylabel={Accuracy [\%]},
ymajorgrids,
ymin=0, ymax=100,
]
\addplot [thick, redisch, dotted, mark=*, mark size=1.5, mark options={solid}]
table [row sep=\\]{%
1 37.83\\
2 51.73\\
3 61.1\\
4 66.99\\
5 70.93\\
6 71.59\\
7 73.01\\
8 73.84\\
};
\addlegendentry{Cit.~\cite{Sundaram2019}}
\addplot [thick, greyisch, dotted, mark=*, mark size=1.5, mark options={solid}]
table[row sep=\\]{%
1 35.75\\
2 48.83\\
3 57.8\\
4 61.72\\
5 65.24\\
6 68.81\\
7 70.2\\
8 73.63\\
};
\addlegendentry{Ours}
\end{axis}

\end{tikzpicture}

  }

  \caption{Comparison of the accuracy on MIT-STAG dataset.}
  \label{fig:comparison_acc}
\end{figure}
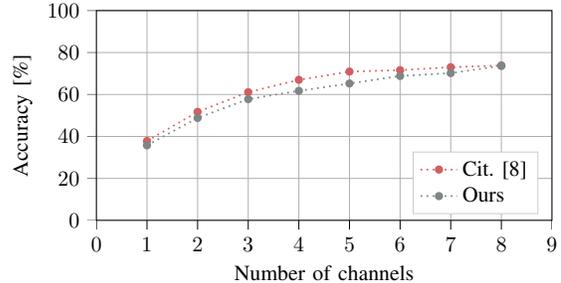

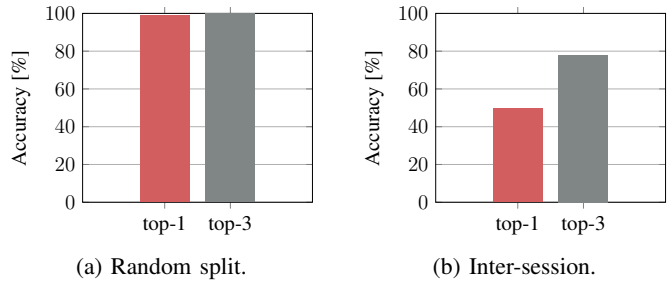
\begin{figure}
\centering
    \begin{subfigure}{0.47\columnwidth}
  \resizebox{\columnwidth}{!}{%
    \centering
\Large
\begin{tikzpicture}
\begin{axis}[
         ybar,
         xmin=0.8,xmax=2.2, 
         ymin=0,
         ymax=100,
         ymajorgrids,
         y grid style={white!69.01960784313725!black},
         area legend,
         xtick={1.3,1.7}, 
         xticklabels={top-1, top-3}, 
         every axis plot/.append style={ 
          bar width=.3,
          bar shift=0pt,
          fill},
          ylabel={Accuracy [\%]},
         ]
         \addplot[draw=none,fill=redisch] coordinates {(1.3,98.86)}; 
         \addplot[draw=none,fill=greyisch] coordinates {(1.7,99.83)};
       \end{axis}
            \end{tikzpicture}
            
}

    \caption{Random split.}
  \label{fig:randomsplit}
\end{subfigure}
\hfill
\begin{subfigure}{0.47\columnwidth}
  \resizebox{\columnwidth}{!}{%
    \centering
\Large
\begin{tikzpicture}
\begin{axis}[
         ybar,
         xmin=0.8,xmax=2.2, 
         ymin=0,
         ymax=100,
         ymajorgrids,
         y grid style={white!69.01960784313725!black},
         area legend,
         xtick={1.3,1.7}, 
         xticklabels={top-1, top-3}, 
         every axis plot/.append style={ 
          bar width=.3,
          bar shift=0pt,
          fill},
          ylabel={Accuracy [\%]},
         ]
         \addplot[draw=none,fill=redisch] coordinates {(1.3,49.63)};
         \addplot[draw=none,fill=greyisch] coordinates {(1.7,77.84)};
       \end{axis}
            \end{tikzpicture}
            
}

    \caption{Inter-session.}
  \label{fig:intersession}
\end{subfigure}
\caption{Classification accuracy on ETHZ-STAG dataset.}
    \label{fig:accuracy}
\end{figure}

